%% file: main.tex
\documentclass[letterpaper]{article} 
\usepackage{aaai24}  
\usepackage{times}  
\usepackage{helvet}  
\usepackage{courier}  
\usepackage[hyphens]{url}  
\usepackage{graphicx} 
\usepackage{natbib}  
\usepackage{caption} 
\usepackage{algorithm}
\usepackage{algorithmic}

\usepackage{xspace} 
\usepackage{bm}
\usepackage{threeparttable}
\usepackage{algorithm}
\usepackage{algorithmic, graphicx}
\usepackage{amsmath,amsthm}
\usepackage{mathrsfs}
\usepackage{booktabs}
\usepackage{subfigure}
\usepackage{color}
\usepackage{rotating}
\usepackage{amsfonts}       
\usepackage{nicefrac}       
\usepackage{bm}
\usepackage{threeparttable}
\usepackage{mathrsfs}
\usepackage{lipsum}
\usepackage{amsthm} 
\usepackage{multirow}
\usepackage{newfloat}
\usepackage{listings}

\usepackage{newfloat}
\usepackage{listings}
\DeclareCaptionStyle{ruled}{labelfont=normalfont,labelsep=colon,strut=off} 
\floatstyle{ruled}
\newfloat{listing}{tb}{lst}{}
\floatname{listing}{Listing}
\title{Learning to Optimize Permutation Flow Shop Scheduling via Graph-based Imitation Learning}
\author{
    Longkang Li\textsuperscript{\rm 1,2}, Siyuan Liang\textsuperscript{\rm 3}, Zihao Zhu\textsuperscript{\rm 1}, Chris Ding\textsuperscript{\rm 1}, Hongyuan Zha\textsuperscript{\rm 1}, Baoyuan Wu\textsuperscript{\rm 1}\thanks{Corresponding to Baoyuan Wu (wubaoyuan@cuhk.edu.cn).}
}
\affiliations{
    \textsuperscript{\rm 1}The School of Data Science, The Chinese University of Hong Kong, Shenzhen (CUHK-Shenzhen), China\\
    \textsuperscript{\rm 2}Mohamed bin Zayed University of Artificial Intelligence, UAE\\
    \textsuperscript{\rm 3}National University of Singapore, Singapore\\

}

\usepackage{bibentry}

\begin{document}

\maketitle

\begin{abstract}
The permutation flow shop scheduling (PFSS), aiming at finding the optimal permutation of jobs, is widely used in manufacturing systems. When solving large-scale PFSS problems, traditional optimization algorithms such as heuristics could hardly meet the demands of both solution accuracy and computational efficiency, thus learning-based methods have recently garnered more attention. Some work attempts to solve the problems by reinforcement learning methods, which suffer from slow convergence issues during training and are still not accurate enough regarding the solutions. To that end, we propose to train the model via expert-driven imitation learning, which accelerates convergence more stably and accurately. Moreover, in order to extract better feature representations of input jobs, we incorporate the graph structure as the encoder. The extensive experiments reveal that our proposed model obtains significant promotion and presents excellent generalizability in large-scale problems with up to 1000 jobs. Compared to the state-of-the-art reinforcement learning method, our model's network parameters are reduced to only 37\% of theirs, and the solution gap of our model towards the expert solutions decreases from 6.8\% to 1.3\% on average. The code is available at: \url{https://github.com/longkangli/PFSS-IL}.
\end{abstract}

\input{main_content.tex}

\section*{Acknowledgments}

Baoyuan Wu is supported by the National Natural Science Foundation of China under grant No.62076213, Shenzhen Science and Technology Program under grant No.RCYX20210609103057050, No.GXWD20201231105722002-20200901175001001, ZDSYS20211021111415025, and
the Guangdong Provincial Key Laboratory of Big Data Computing, the Chinese University of Hong Kong, Shenzhen.

\bibliography{mybib}

\input{appendix}

\end{document}

%% file: main_content.tex
\section{Introduction}
Scheduling is a synthesized process of arranging, controlling, and optimizing jobs and workload in a manufacturing system. Here we utilize machine learning to solve a frequently encountered optimization problem: the permutation flow shop scheduling (PFSS) \cite{sharma2021improved}, which is about sequentially processing several jobs on a series of machines. The PFSS is one of the most comprehensively studied scheduling problems, with broad applications in production and planning \cite{ribas2010review}, intuitive designs \cite{alfaro2020automatic}, and transportation \cite{soukhal2005complexity}. 

\begin{figure}[!ht]
	\centering
	\setlength{\abovecaptionskip}{0.15cm}
	\subfigure[The overview of the PFSS.]{
		\centering
		\includegraphics[width=0.9\linewidth]{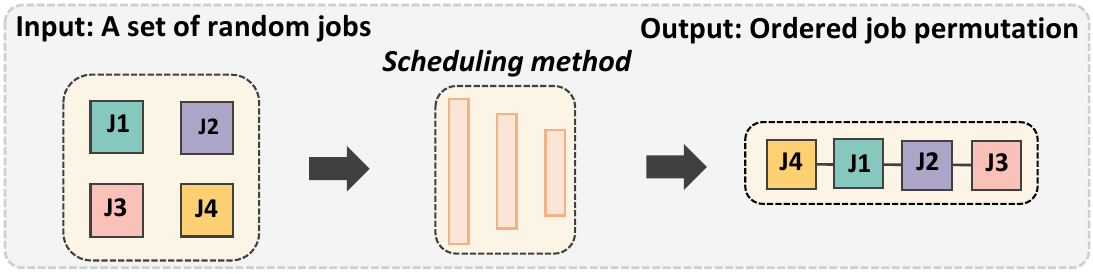}
		\label{fig:1_a}
	}\\
	\subfigure[Gantt charts of different permutations.]{
		\centering
		\includegraphics[width=0.9\linewidth]{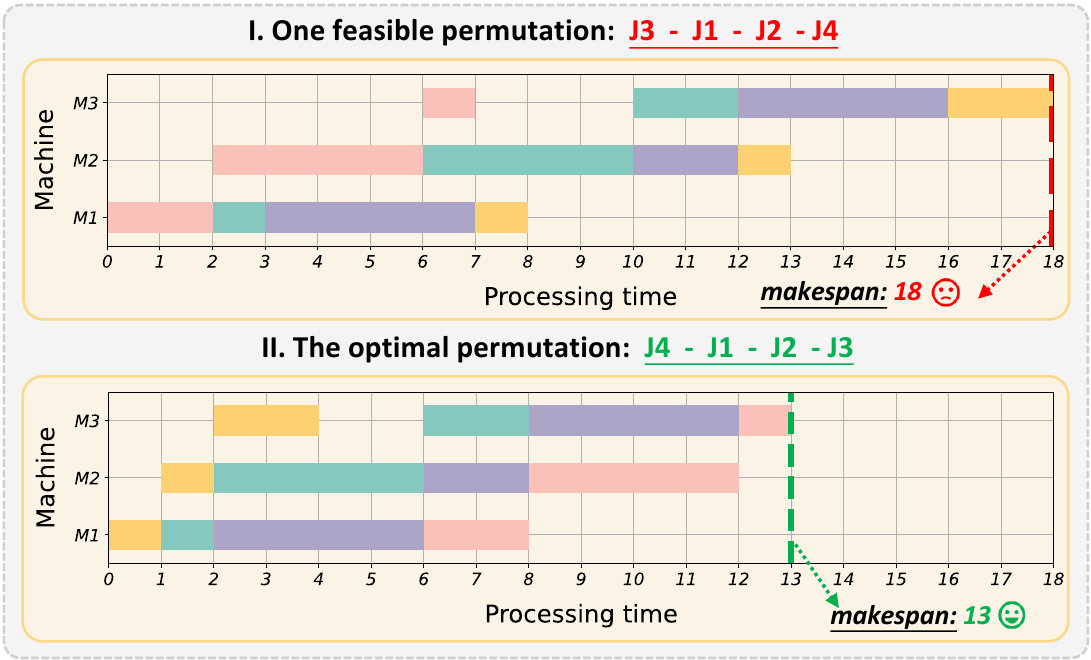}
		\label{fig:1_b}
	}
	\caption{\small{An overview of a $4$-job (J$1${$ \sim $}J$4$) $3$-machine (M$1${$ \sim $}M$3$) PFSS problem. One color corresponds to one job. All jobs follow the same operational order M1-M2-M3. \textbf{(a)} A set of jobs is scheduled into an ordered job permutation via a scheduling method. The scheduling method aims at finding the optimal job permutation. \textbf{(b)} The comparison between two permutations on the \textit{Gantt charts}:  Given the same input set of jobs, different job permutation has different makespan. The optimal permutation minimizes the makespan. Refer to Section \ref{sec_problem} for more properties of the PFSS.}}
	\label{fig1}
	\vspace{-1em}
\end{figure}


Figure \ref{fig1} gives an example and overview of a $4$-job $3$-machine PFSS problem. One machine corresponds to one operation, and the PFSS strictly requires all the jobs to be processed in the same order of operations. Typically, the goal of the PFSS is to find the optimal permutation of the jobs, which minimizes the makespan. Makespan describes the total processing time from the start to the end for scheduling a sequence of jobs. 
When there is only $1$ machine, this PFSS problem can be trivially solved. As for a $n$-job $m$-machine PFSS problem ($m\geq2$), the number of feasible solutions should be $n!$. The PFSS thus can be considered as a combinatorial optimization (CO) problem with the NP-hard property \cite{lenstra1981complexity}.

Recently, machine learning has been widely used in combinatorial optimization problems \cite{bengio2020machine, li2022bilevel}. One popular model is pointer network (PN) \cite{vinyals2015pointer}, which uses the Long Short Term Memory (LSTM) as an encoder and the attention mechanism as a decoder (or pointer) to select a member of the input as the output. The PN can generalize a small pre-trained model to arbitrarily large instances. The current state-of-the-art learning-based method for solving the PFSS utilizes the PN and training the network via actor-critic as a paradigm of reinforcement learning (RL) \cite{pan2020solving}. Besides the learning-based methods, there are traditional optimization algorithms for solving the PFSS, including the mathematical models and the heuristics. Mathematical models \cite{rios1998computational} are exact methods that are guaranteed to obtain optimal solutions. However, it takes a tremendous amount of time to solve due to the NP-hardness of the PFSS. The heuristics \cite{sharma2021improved} allow getting a feasible or sometimes quasi-optimal solution within a limited time but still encounter the trade-off between the solution effectiveness and computational efficiency. 

\textbf{Motivations.} For solving large-scale PFSS problems in practical scenarios, traditional optimization algorithms could hardly satisfy the demands of both solution accuracy and computational efficiency, thus the learning-based methods attract more attention. However, the state-of-the-art RL method \cite{pan2020solving} requires heavy networks (an actor-network and a critic network) and takes a long time for training until the network converges. To that end, we train the model via expert-driven imitation learning (IL) \cite{monteiro2020augmented}, which leads to more stable and accurate convergences. Moreover, in order to extract better feature representations of input jobs, we incorporate the graph structure, the Gated Graph ConvNets (GGCN) \cite{bresson2018experimental}, for obtaining better job feature representations. More technical motivations for using graph encoder and attention decoder, and how our model pre-trained on a small-sized job can be generalized to arbitrarily larger-sized jobs, will be explained in Section \ref{method}.

To sum up, we make the following contributions in this paper: \textbf{1)} To the best of our knowledge, we are the first to solve the PFSS problem via expert-driven imitation learning, which accelerates convergence faster and more efficiently. \textbf{2)} Our learning model is based on the graph structure, which achieves a better representation capability. Our graph-based imitation learning model is much lighter in parameter weights, more stable in convergence, and more accurate in performance, compared to the state-of-the-art reinforcement learning methods. \textbf{3)} We explore and extend the job number to $1000$ in solving PFSS problems. Extensive experiments on both benchmark datasets and generated datasets show the competitive performance of our proposed model, especially the excellent generalizability. 


\section{Related work}


\textbf{Traditional optimization algorithms for PFSS.} \ Traditional optimization algorithms for solving the PFSS include mathematical models and heuristics. The PFSS can be formulated as mixed-integer programming. Many different mathematical models have been proposed for the PFSS and solved by the exact methods such as branch-and-bound and brand-and-cut methods \cite{rios1998computational}. Over the last few decades, many heuristics have been applied to the PFSS problems. Random search \cite{zabinsky2009random} can settle ill-structured global optimization problems with continuous and discrete variables. Iterated local search \cite{lourencco2019iterated} is based on a local search strategy using a single solution along the iterative process. The iterated greedy \cite{ruiz2019iterated} consists of improved initialization, iterative construction, and destruction procedures, along with a local search. Sharma et al. \cite{sharma2021improved} apply a tie-breaking rule to the Nawaz-Enscore-Ham (NEH) heuristic. For the large-scale PFSS problems, the heuristics could still hardly balance the solution accuracy and runtime efficiency: random search and iterated local search obtain poor makespans, while iterated greedy and NEH algorithms are time-consuming.  


\noindent\textbf{Learning-based methods for PFSS.} \ With the development of deep learning (DL) and deep reinforcement learning (DRL) \cite{bengio2020machine, li2022learning}, some work has been starting to solve the scheduling problems from the RL perspectives. Ren et al. used value-based methods to solve the PFSS, such as Sarsa and Q-learning \cite{ren2021solving}. However, this method focuses on the scale of less than 100 jobs and performs poorly in large-scale problems. R. Pan et al. \cite{pan2020solving} applied PN \cite{vinyals2015pointer} in solving the PFSS and explored the actor-critic to train the PFSS models, which achieved the state-of-the-art performances for solving the PFSS. Z. Pan et al. \cite{pan2020} also used actor-critic to train the model. The difference was that \cite{pan2020} still used recurrent neural network (RNN) as an encoder whereas \cite{pan2020solving} used LSTM as an encoder which incorporated the gating mechanism. However, those RL-based methods take a long time for the network training until the convergence, and the solution accuracy still needs more improvements. Our experiments mainly compared with \cite{pan2020solving}. For more related works (such as learning-based methods for combinatorial optimization, and imitation learning), please refer to \textit{Appendix A}.

\section{Problem descriptions}
\label{sec_problem}
 As shown in Figure \ref{fig1}(a), the goal of the PFSS is to output the optimal permutation of the jobs, which minimizes the makespan. We denote the number of jobs as $n$ and the number of machines as $m$ ($m{\geq}2$). Let $i$ be the machine index, $j$ be the job index, $i \in I$=$\{1,2,...,m\}$, $j \in J$=$\{1,2,...,n\}$. Then the inputs and outputs for the PFSS are given as:



\begin{itemize}
    \item \textit{Inputs}: matrix of processing times $\boldsymbol{X}_{m\times n}$. Each element $x_{ij}$ denotes the processing time of job $j$ on machine $i$, $x_{ij} {\geq} 0, \  x_{ij} {\in} \boldsymbol{X}_{m\times n}$; 
    
    \item \textit{Outputs}: the optimal permutation $\boldsymbol{\tau}^{*}$=$[\tau_0^{*},\tau_1^{*},...,\tau_{n-1}^{*}]$, where the corresponding makespan is minimized. 
    
\end{itemize}

Figure \ref{fig1}(b) utilizes the Gantt charts to reveal the scheduling process of different job permutations. Givens the same inputs $\boldsymbol{X}_{m\times n}$, different job permutation leads to different makespan. Herein, the PFSS process can be characterized by the following properties:
\begin{itemize}
    \item It is assumed that the start time is zero.
    \item One machine corresponds to only one operation. One machine can only process one job at a time.
    \item Each job needs to be processed on every machine once, with the pre-defined operational order.
    \item All the jobs share the same operational order.
    \item Once started, the operation can not be interrupted. 
\end{itemize}

According to \cite{rios1998computational}, the PFSS can be mathematically formulated as a mixed-integer programming model, which is given in the \textit{Appendix B}.

\section{Proposed method}
\label{method}

In this Section, we formulate the PFSS as a Markov decision process (MDP) \cite{howard1960dynamic} in Section 4.1, and demonstrate our encoder-decoder policy network in Section 4.2. The policy network training via the IL is shown in Section 4.3. The comparison between the LSTM encoder and our GGCN encoder is exhibited in Figure \ref{rnn}. The training overview of our proposed IL model is given in Figure \ref{fig3}.       

\subsection{4.1. Markov decision process for PFSS}
\label{mdp}
\begin{figure}[t!] 
    \centering 
    \setlength{\abovecaptionskip}{0.2cm}
    \includegraphics[width=0.4\textwidth]{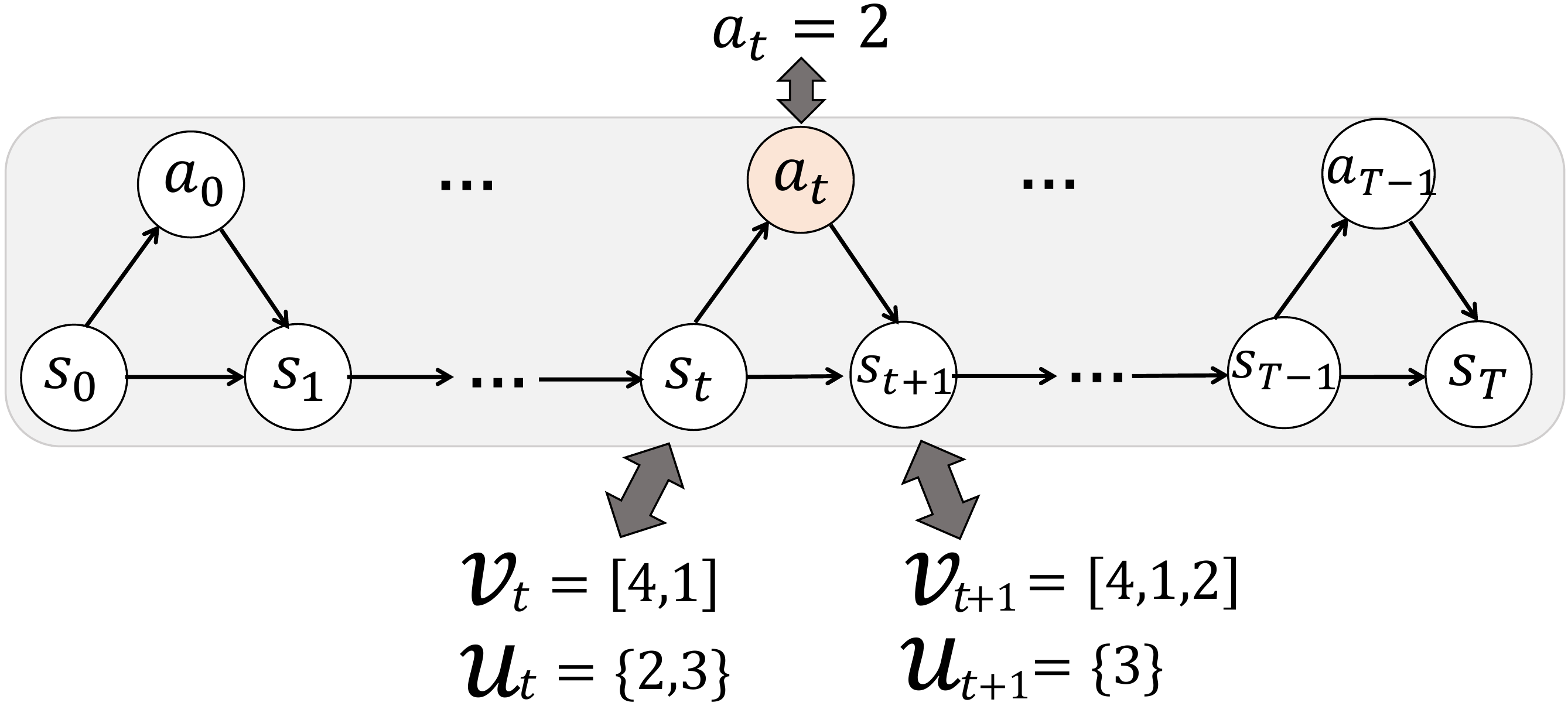}
    \caption{An example of the PFSS as a MDP: the state $s_t$ transits to state $s_{t+1}$ through the action $a_t$, which is obtained via the policy $\pi_\theta(\textbf{a}|s_t)$. Each state is a concatenation of scheduled job list $\bm{\mathcal{V}}$ and unscheduled job set $\bm{\mathcal{U}}$. The figure is scheduling $T{=}n{=}4$ jobs with final output $\bm{\tau}=[4,1,2,3]$.}
    \label{fig2} 
    \vspace{-1em}
\end{figure}

\begin{figure*}[ht] 
    \centering 
    \includegraphics[width=0.83\textwidth]{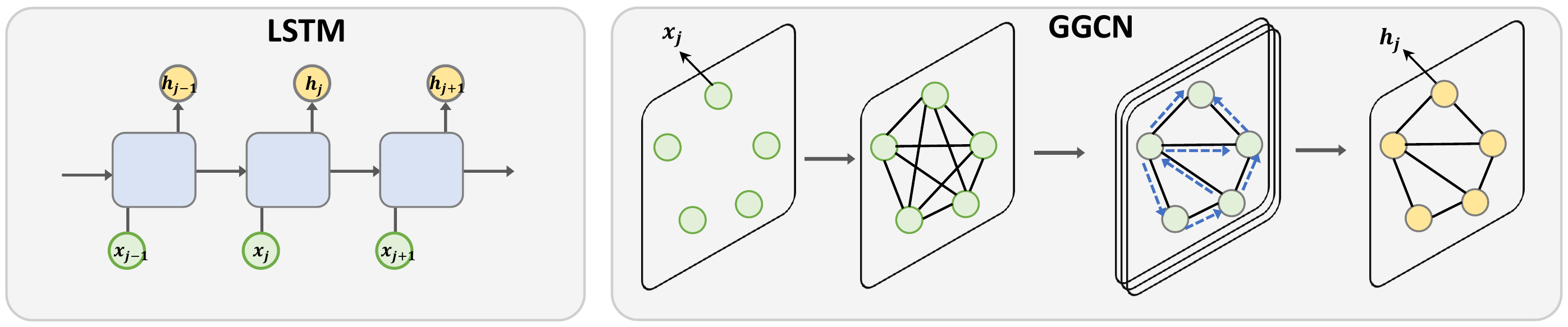}
    \setlength{\abovecaptionskip}{0cm}
    \caption{The comparison between the two encoders. \textbf{Left:} recurrence-based encoder inputs each job features $\boldsymbol{x}_j$ (green) and outputs their job embeddings $\boldsymbol{h}_j$ (yellow). \textbf{Right:} graph-based encoder inputs the job features $\boldsymbol{x}_j$, firstly construct a fully-connected graph, then derive a sparse graph based on sparsification techniques, and finally get each job embedding $\boldsymbol{h}_j$ after $\mathcal{L}$ layers graph operations. $j \in \{1,2,...,n\}$.}
    \label{rnn} 
\end{figure*}

\begin{figure*}[ht] 
    \centering 
    \includegraphics[width=0.85\textwidth]{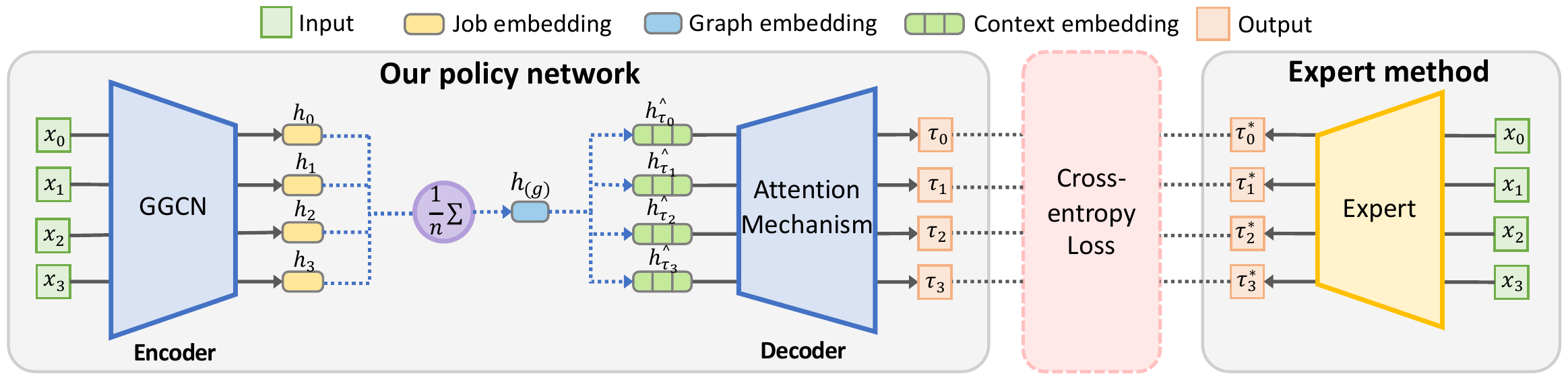}
    \setlength{\abovecaptionskip}{0cm}
    \caption{The training overview of our proposed IL method for solving $n$-job $m$-machine PFSS. \textbf{Left:} our proposed encoder-decoder policy network. \textbf{Right:} obtaining the expert solutions for imitation learning.  \textbf{Middle:} The policy network is trained by minimizing the cross-entropy loss. When testing, we simply go through the left part to obtain the outputs. $n{=}4$ in the figure.}
    \label{fig3} 
\end{figure*}

The PFSS problem is about scheduling for a permutation of $n$ jobs. If scheduling one job is a one-step decision, then the sequential decisions of totally $n$ steps made for scheduling one $n$-job PFSS problem can be seen as MDP, as shown in Figure \ref{fig2}. Considering the PFSS-solving process as the environment and the scheduling method as the agent, we set the state, action, and policy as:       

\begin{itemize}
    \item \textit{State} $s_t$: the current state of the scheduling, represented by [$\bm{\mathcal{V}_t}$, $\bm{\mathcal{U}_t}$], meaning the concatenation of the already scheduled job list and the remaining unscheduled job set. $t \in \{0,1,2,...,n\}$.
    \item \textit{Action} $a_t$: a job index, chosen to process next from the unscheduled job set. The action is obtained by the policy network with masking, which enforces that the selected job as action $a_t$ is always from the unscheduled set $\bm{\mathcal{U}_t}$. The masking mechanism can help to avoid selecting the scheduled job again. $t \in \{0,1,...,n{-}1\}$.
    \item \textit{Policy} $\pi_\theta(\textbf{a}|s_t)$: the policy $\pi$ determines how the actions proceed, where $\theta$ represents the weights of the network. It outputs a probability distribution over all jobs, denoted by $\textit{\textbf{p}} = \pi_\theta(\textbf{a}|s_t)$. \textbf{a} is the set of actions.   
\end{itemize}

Scheduling a $n$-job PFSS problem can be seen as a $n$-step sequential decision process. The initial state $s_0$ is that all jobs are unscheduled, and the final state $s_n$ is that all job are scheduled to obtain the permutation of $n$ jobs, denoted by $\bm{\tau} {=} [\tau_0,\tau_1,...,\tau_{n-1}] \in \mathcal{T}$, where $\tau_t {=} a_t$, $t {\in} \{0,1,2,...,n{-}1\}$. 


\subsection{4.2. Policy network}
\label{policy}

We follow the encoder-decoder structure of previous work for the PFSS problems \cite{pan2020solving}. We incorporate the graph structure GGCN \cite{bresson2018experimental} as an encoder and follow previous work to utilize the attention mechanism as the decoder. Particularly, the motivations and the technical details for the encoder and decoder are given in the following from \textbf{1)} to \textbf{4)}. We explain how our model pre-trained on small-sized jobs can be generalized to larger-sized jobs in \textbf{5)}.


\vspace{0.2em}
\noindent\textbf{1) Motivations of using graph encoder}. The goal of our PFSS task is to find the optimal job permutation which minimizes the makespan. The input is $\boldsymbol{X}_{m\times n}$, where $j$-th column vector $\boldsymbol{x}_j {\in} \mathbb{R}^{m}$ denotes the feature for job $j$, $j$=$\{1,..,n\}$. If all jobs have the same features, the problem could be trivially solved, because any permutation leads to the same makespan; Otherwise, different permutations correspond to different makespans. Thus we believe that the difference between job features could impact on the permutation results. Based on this observation, we incorporate the graph structure, regarding one job feature as one node and using the edge to represent the difference between two jobs. As mentioned in Eq.(\ref{eq1}), we initialize the edge by the embedding of the Euclidean distance between two adjacent jobs. In our experiments, we evaluate how job difference influences the permutation results in Section 5.2.(5) and Figure \ref{fig5}(c). Actually, the results proved our assumption. Compared to the previous recurrence-based LSTM encoder, our graph-based GGCN encoder explicitly formulates the embeddings of job and job differences using the node and edge modules, respectively. 

\vspace{0.2em}
\noindent\textbf{2) Graph encoder}. In the $n$-job $m$-machine PFSS problem, the input is a matrix $\boldsymbol{X}_{m\times n}$ of all processing times, where $x_{ij} \in \boldsymbol{X}_{m\times n}$ represents the processing time of job $j$ on machine $i$. We denote $\bm{x}_j^T{=}[x_{1j},x_{2j},...,x_{mj}]$ as the input features of job $j$. 
As shown in Figure \ref{rnn}, the encoder starts with a fully connected graph, and derives a sparse graph using sparsification techniques. We follow Dai et. al. \cite{dai2017learning} to use a fixed graph diameter and connect each node in the graph to its $n {\times} 20\%$-nearest neighbors by default.   
Anisotropic GNNs such as GGCN \cite{bresson2018experimental} have shown competitive performances across several challenging domains \cite{joshi2021learning}. In this paper, we choose GGCN as our encoder. Let $\bm{h}_j^\ell$ and $\bm{e}_{jk}^\ell$ denote the job and edge feature at layer $\ell$ associated with job $j$ and edge $jk$, $\ell \in \{0,1,...,\mathcal{L}{-}1\}$, 

\vspace{-0.8em}
\begin{footnotesize}
\begin{align}
    \label{eq1}
    \begin{cases}
    \bm{h}_j^0 = W^h \bm{x}_j, \ \bm{h}_k^0 = W^h \bm{x}_k, \ \bm{e}_{jk}^0 = W^e \cdot \Vert \bm{x}_j - \bm{x}_k \Vert_2   \\
    \bm{h}_j^{\ell+1}\! = \!\bm{h}_j^{\ell} \!+\! {\rm{ReLU}}\!\left({\rm{Nm}}\!(B^{\ell}\bm{h}_j^{\ell}+{\rm{Ag}}_{k\in \bm{N}_j}(\sigma(\bm{e}_{jk}^{\ell}) \odot C^{\ell}\bm{h}_k^{\ell}))\right) \\
    \bm{e}_{jk}^{\ell+1} \!=\! \bm{e}_{jk}^{\ell}\! +\! {\rm{ReLU}}\!\left({\rm{Nm}}\!\left(D^{\ell}\bm{e}_{jk}^{\ell} + E^{\ell}\bm{h}_j^{\ell} + F^{\ell}\bm{h}_k^{\ell} \right)\right),  
    \end{cases}
\end{align}
\end{footnotesize}
\vspace{-0.3em}

\noindent where $\bm{x}_j, \bm{x}_k \in \mathbb{R}^{m}$ are the input job features, $\bm{h}_j^0$, $\bm{h}_k^0 \in \mathbb{R}^{d}$ are their linear projections via $W^h {\in} \mathbb{R}^{d\times m}$. $\bm{e}_{jk}^0 \in \mathbb{R}^{d}$ is the embedding of the Euclidean distance of $j$ and $k$ via $W^e{\in} \mathbb{R}^{d}$. And $B^\ell, C^\ell, D^\ell, E^\ell, F^\ell \in \mathbb{R}^{d\times d}$ are the parameters to learn. Nm($\cdot$) denotes the method of normalization: LayerNorm \cite{ba2016layer} or BatchNorm \cite{ioffe2015batch}. Ag($\cdot$) means the aggregation functions: SUM, MEAN or MAX. $\sigma$ is the sigmoid function, and $\odot$ is the Hadamard product. We make the aggregation function anisotropic using the edge gates $\sigma(\bm{e}_{jk})$ via dense attention mechanism which scales the neighborhood features $\bm{h}_k, \forall k {\in} \bm{N}_j {=}\{0,...,n{-}1\}\backslash \{j\}$, $ \bm{N}_j$ is the neighbor set of $j$.  


\vspace{0.2em}
\noindent\textbf{3) Motivations of using attention decoder}. The PFSS problems actually have feasibility constraints. When decoding, previously scheduled jobs cannot be selected next. Thus, the earlier scheduled jobs have more impact on the overall makespan than the latter jobs. It is not technically enough if the decoder just uses a simple masking procedure as Eq.(\ref{eq4}). In essence, the attention mechanism allows to focus more on a certain part. Thus we apply the attention, aiming to focus more on the front jobs, to improve the solution quality regarding the overall makespan.

\vspace{0.2em}
\noindent\textbf{4) Attention decoder}. We follow the previous RL work \cite{pan2020solving} to use the attention mechanism \cite{vaswani2017attention} in our decoder. Decoding can be seen as a $n$-step sequential decision-making process. For the $n$-job PFSS problem, at time-step $t \in \{0,1,...,n{-}1\}$, the state of the scheduling process can be denoted as $s_t$, and according to the policy $\pi_\theta$, the decoder will output a selected job to schedule next, denoted as action $a_t$.

\input{table}

There are $\mathcal{L}$ layers in the encoder, and the graph embedding is $\bm{h}_{(g)}{=}\frac{1}{n} \sum_{k=0}^{n-1} \bm{h}_k^{\mathcal{L}}$, where $\bm{h}_k^{\mathcal{L}}$ is the job embedding. In Figure \ref{fig3}, we omit the superscript ${\mathcal{L}}$ for readability.   
At time-step $t$, the decoder constructs a context embedding $\hat{\bm{h}}_{(c)}$ using the encoder and the partial output of the decoder, 

\vspace{-1em}
\begin{align}
  \label{eq2}
  \hat{\bm{h}}_{(c)} {=} \hat{\bm{h}}_{\tau_t} {=}
  \begin{cases}
  [\bm{h}_{(g)}, \bm{h}_{\tau_{0}}^{\mathcal{L}}, \bm{h}_{\tau_{t{-}1}}^{\mathcal{L}}] \ \ t=\{1,...,n{-}1\}\\
  [\bm{h}_{(g)}, \bm{v}, \bm{v}]   \qquad  \ \ \ \  t=0,  \\
   \end{cases}
\end{align}

\noindent where the $\bm{v}$ is the placeholders for the first time decoding, $\bm{h}_{(g)}$ is the graph embedding, $\bm{h}_{\tau_{0}}^{\mathcal{L}}$ is the first job embedding, $\bm{h}_{\tau_{t-1}}^{\mathcal{L}}$ is the previous job embedding, and [$\cdot$,$\cdot$,$\cdot$] is the horizontal concatenation operator. The context embedding $\hat{\bm{h}}_{(c)}$ can be interpreted as the (3$\cdot d$)-dimensional concatenated job embedding, denoted by $\hat{\bm{h}}_{(c)}{=}\hat{\bm{h}}_{\tau_t}$. Next, one more layer of standard MHA refines the context embedding $\hat{\bm{h}}_{(c)}$ to get $\bm{h}_{(c)}$, 

\begin{small}
\begin{align}
  \label{eq3}
 \bm{h}_{(c)} {=} {\rm{MHA}}\left( Q{=}\hat{\bm{h}}_{(c)}, K{=}\{\bm{h}_0^{\mathcal{L}},..,\bm{h}_{n{-}1}^{\mathcal{L}}\}, V{=}\{\bm{h}_0^{\mathcal{L}},..,\bm{h}_{n{-}1}^{\mathcal{L}}\} \right).
 \end{align}
\end{small}
 
\noindent The MHA uses $M{=}8$ heads, with input $Q, K, V$. The logits $u_{(c)j}$ for the edge $\bm{e}_{(c)j}$ are obtained via a final single-head attention,
 

\vspace{-1em}
\begin{align}
  \label{eq4}
  u_{(c)j} {=} 
  \begin{cases}
  r \cdot {\rm{tanh}}\left(\frac{(W^Q \bm{h}_{(c)})^T\cdot(W^K \bm{h}_j^{\mathcal{L}})}{\sqrt{d}}\right) \  {\rm{if}} \  j \in \bm{\mathcal{U}}_t \\
  -\infty  \qquad \qquad \qquad \qquad \qquad \quad   \rm{otherwise}, 
  \end{cases}
\end{align}

\noindent where the tanh function is used to maintain the value of logits within [-$r$, $r$] ($r$=$10$) \cite{bello2016neural}, $W^Q$ and $W^K$ are the parameters to learn, $\bm{\mathcal{U}}_t$ is the current unscheduled job set. The masking enforces that the scheduled jobs will not be selected again by setting the logits of those scheduled jobs to $-\infty$. We calculate the final output probability vector $\textit{\textbf{p}}$ using a softmax, each element $p_j$ can be obtained by, 

\vspace{-1em}
\begin{align}
  \label{eq5}
  {p_j} = \frac{e^{u_{(c)j}}}{\sum_k e^{u_{(c)k}}}.
\end{align}

\noindent According to the output probability vector $\textit{\textbf{p}}$, we can select the job whose probability is the highest to process next, denoted by $\tau_t$ or $a_t$. Update the scheduled job list $\bm{\mathcal{V}}$ by appending $a_t$, and update the unscheduled set $\bm{\mathcal{U}}$ by eliminating $a_t$. 

From Eq.\eqref{eq2} to \eqref{eq5}, after taking all the $n$ steps for the decoding process, we will obtain the final output permutation of all jobs, denoted by $\bm{\tau} {=} (\tau_0,\tau_1,...,\tau_{n-1}) \in \mathcal{\bm{T}}$.

\vspace{0.2em}
\noindent\textbf{5) Availability of generalization.} We would like to explain how our model can be pre-trained only on small job sizes and then generalized to any other job size. We can briefly interpret the encoding and decoding process by the projection: $\boldsymbol{H}_{d\times n}$ = $W^{en}_{d\times m}$ $\boldsymbol{X}_{m\times n}$ and $\boldsymbol{O}_{1\times n}$ = $W^{de}_{1\times d}$ $\boldsymbol{H}_{d\times n}$, where $\boldsymbol{X},\boldsymbol{H},\boldsymbol{O},d$ refer to the input, hidden representation, output and hidden dimension. $W^{en}$ and $W^{de}$ denote the network weights for the encoder and decoder, respectively. As mentioned above, one step of decoding obtains one job and we need to go through $n$ steps to obtain the permutation of $n$ jobs. It is obvious that the dimensions of $W^{en}_{d\times m}$ and $W^{de}_{1\times d}$ are only relevant to $m$ and $d$, not to $n$. To that end, it enables the generalizability to different job sizes, though pre-trained on small job sizes. However, their machine number $m$ for training and testing sets should always be consistent.

\subsection{4.3. Expert-driven imitation learning}
\label{il}
\textbf{Motivations.} RL is a currently popular approach to finding good scheduling policies, such as \cite{pan2020solving}. However, RL methods train from scratch and possibly run into many issues: the randomly initialized policies may perform poorly, and the convergence could be slow in training. In this paper, instead of using RL, we choose to learn directly from an expert method, referred to as IL \cite{hussein2017imitation}. 

During our model training using IL, we chose the state-of-the-art NEH heuristic \cite{sharma2021improved} as the expert method. The mathematical model as formulated in the \textit{Appendix 1} solved by branch-and-bound method \cite{gmys2020computationally} is another potential option as the expert method, due to its exact solutions. However, the computational time is much slower than expected, and the processing times of inputs require only to be integers. Considering these limitations, we would rather choose the NEH heuristic as the expert method. More experimental analysis is given in \textit{Appendix C}.



We choose behavior cloning \cite{torabi2018behavioral} as the training method of IL. We firstly run the expert on a set of training instances and record a dataset of expert state-action pairs $\mathcal{R} {=} \{(s^i_t,{a}^{*i}_t )_{t=0}^{n-1}\}_{i{=}0}^{N_1{-}1}$, where $N_1$ is the number of training instances, and each instance is a $n$-job $m$-machine PFSS problem. The set of actions by the expert constructs a permutation of job: $\bm{\tau}^{*}$=$\{\tau_0^{*},\tau_1^{*},...,\tau_{n-1}^{*}\}$=$\{{a}_{0}^{*},{a}_{1}^{*},...,{a}_{n-1}^{*}\}$. The policy network is learned by minimizing the cross-entropy loss,

\vspace{-1em}
\begin{align}
  \label{eq15}
  \mathcal{J}(\theta) = - \frac{1}{N_1{\cdot}n} \sum_{i=0}^{N_1{-}1} \sum_{t=0}^{n{-}1} \log\pi_\theta({a}_t^{*i}|s_t^i).
\end{align}
\vspace{-1em}

\section{Experiments}
\label{exp}

\subsection{5.1. Setup}
\label{setup}

\input{benchmark2}

\begin{table*}[!t]
    \centering
    \setlength{\abovecaptionskip}{0cm}
    \caption{Time complexity analysis on different methods. The total time complexities come from: getting expert time (only for our IL method), training time (only for learning-based methods), and testing time. Refer to Subsection "Implementation details" and "Time complexity analysis" below for more details of the notations.}
    \label{tab5}
    \resizebox{17.5cm}{!}{ 
    \begin{tabular}{ccccc}
    \toprule[1.5pt]
    Method & Getting Expert & Training & Testing & \textbf{Total}  \\
    \midrule
    NEH \cite{sharma2021improved} & / & / & $\mathcal{O}(N_3{\cdot}{m}{n}^3)$ & $\mathcal{O}(N_3{\cdot}{m}{n}^3)$ \\
    RL \cite{pan2020solving} & /
    & $\mathcal{O}({N_1}{\cdot}(m{\hat{n}}{\cdot}F_{RL}{+}m{\hat{n}}{\cdot}B_{RL}))$ 
    & $\mathcal{O}(N_3{\cdot}mn{\cdot}F_{RL})$  
    & $\mathcal{O}(N_1{\cdot}m{\hat{n}}(F_{RL}{+}B_{RL}) + N_3 F_{RL}{\cdot}mn )$ \\
    {IL [Ours]} & $\mathcal{O}(N_1{\cdot}m{\hat{n}}^3)$ 
    & $\mathcal{O}({N_1}{\cdot}(m{\hat{n}}{\cdot}F_{IL}{+}m{\hat{n}}{\cdot}B_{IL}))$ 
    & $\mathcal{O}(N_3{\cdot}mn{\cdot}F_{IL})$  
    & $\mathcal{O}(N_1{\cdot}m{\hat{n}}({\hat{n}}^2{+}F_{IL}{+}B_{IL}) + N_3 F_{IL}{\cdot}mn )$ \\
    \bottomrule[1.5pt]
    \end{tabular} 
    }
\end{table*}

\begin{table*}[!t]
    \centering
    \setlength{\abovecaptionskip}{0cm}
    \caption{\small{Time cost analysis on different methods. We evaluate different datasets, based on the same pre-trained model on $12800$ training instances with $\hat{n}{=}20,m{=}5$. The total time consists of getting expert time, training time, and testing time. The lowest time cost is in \textbf{BOLD}.}}
    \resizebox{18cm}{!}{ 
    \begin{tabular}{ccccccccccccc}
    \toprule[1.5pt]
    \multicolumn{3}{c}{Dataset:} & \multicolumn{2}{c}{VRF Benchmark} & \multicolumn{2}{c}{Taillard Benchmark} & \multicolumn{6}{c}{Generated Dataset} \\
    \midrule

    \multicolumn{3}{c}{Testing size: ($N_3,n,m$)} & \multicolumn{2}{c}{(10,60,5)} & \multicolumn{2}{c}{(10,100,5)} & \multicolumn{2}{c}{(100,20,5)} & \multicolumn{2}{c}{(100,100,5)} &  \multicolumn{2}{c}{(100,1000,5)} \\
    \cmidrule(lr){1-3}\cmidrule(lr){4-5}\cmidrule(lr){6-7}\cmidrule(lr){8-13}
    
    Method & Getting Expert$\downarrow$ & Training$\downarrow$ & Testing$\downarrow$ & {Total}$\downarrow$ & Testing$\downarrow$ & {Total}$\downarrow$ & Testing$\downarrow$ & {Total}$\downarrow$ & Testing$\downarrow$ & {Total}$\downarrow$ & Testing$\downarrow$ & {Total}$\downarrow$ \\
    \cmidrule(lr){1-3}\cmidrule(lr){4-5}\cmidrule(lr){6-7}\cmidrule(lr){8-9}\cmidrule(lr){10-11}\cmidrule(lr){12-13}
    NEH \cite{sharma2021improved} & 0 & 0 & {3.4s} & \textbf{3.4s} & 15.9s & \textbf{15.9s} & \textbf{1.2s} & \textbf{1.2s} & 154.5s & \textbf{0.04h} & 43.1h & 43.1h \\
    RL \cite{pan2020solving} & 0 & 1550.0s & \textbf{3.3s} & 0.4h & \textbf{3.5s} & 0.4h & 3.0s & 0.4h  & \textbf{3.6s} & 0.4h  & \textbf{81.8s} & 0.5h \\
    {IL [Ours]} & 165.1s & 620.0s & 4.8s & 0.2h & 3.6s & 0.2h & 3.3s & 0.2h & 3.7s & {0.2h} & 89.9s & \textbf{0.3h} \\ 
    \bottomrule[1.5pt] 
    \end{tabular}}
    \label{total}
\end{table*}

\noindent\textbf{1) Datasets.} To evaluate the efficacy of our method, we conduct the experiments on two randomly generated datasets and two benchmarks (Taillard \cite{taillard1993benchmarks} and the VRF \cite{vallada2015new}). Considering that in real life jobs are not always evenly and regularly distributed, we thus choose two challenging distributions for random generation, one is Gamma distribution with $k{=}1, \theta{=}2$, and the other is Normal distribution with $\mu{=}6, \sigma{=}6$. Due to space limitations, we only record the results on the Gamma distribution in the paper and put the detailed results on the Normal distribution in the \textit{Appendix D}.

\noindent\textbf{2) Implementation details.} Let $N_1, N_2, N_3$ be the number of training, validation, and testing instances, and $\hat{n}, n, m$ denote the training job size, the testing job size, and the machine size, respectively. $m$ is set to 5 for all phases by default. We train the model on $N_1{=}12800$ instances with $\hat{n}{=}20$, validate on $N_2{=}1000$ instances with the same job size. We generalize the small pre-trained model to any larger job sizes. For the random datasets, we test on $N_3{=}1000$ instances with $n{=}20,50,100$, and test on $N_3{=}100$ instances with $n{=}200,500,1000$, respectively. For the two benchmarks, we use our pre-trained model based on Gamma distribution with $\hat{n}{=}20,m{=}5$ or $m{=}20$, to test different sets of instances, where each testing size has 10 instances.  


For these learning methods, the batch size is set to $128$, and the learning rate is initialized as $1e{-}4$ with $0.96$ per epoch as decay. We train for $50$ epochs. 
The experiments are conducted in Ubuntu $20.04$ LTS 64-bit System with Intel(R) Xeon(R) Silver $4214$ $2.20$GHz $\times 48$ CPU and one NVIDIA GeForce RTX $2080$Ti GPU. All methods are implemented by Python, and the two learning methods use Pytorch.

\noindent\textbf{3) Baselines.} We compare our method with some heuristics and learning-based methods. We consider four of the popular and mostly-used heuristics including random search \cite{zabinsky2009random}, iterated local search \cite{lourencco2019iterated}, iterated greedy \cite{ruiz2019iterated}, and NEH algorithm \cite{sharma2021improved}. We also compare with the state-of-the-art RL-based methods, where actor-critic is used to solve the PFSS ~\cite{pan2020solving}.

\begin{figure*}[t!] 
    \centering 
    \includegraphics[width=0.90\textwidth]{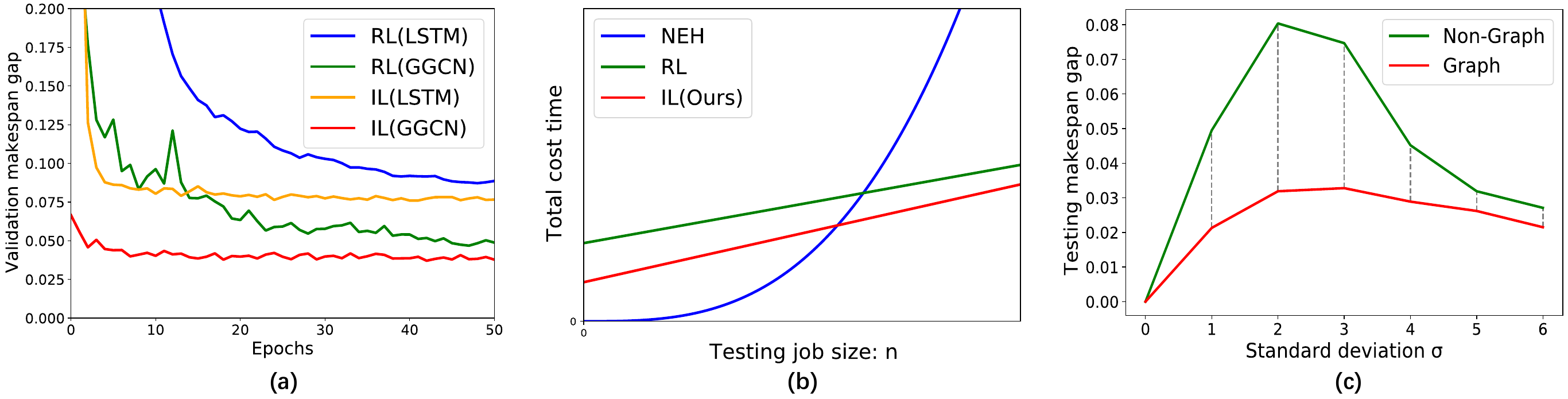} 
    \setlength{\abovecaptionskip}{-0cm}
    \caption{{\textbf{(a)} Ablation study: the validation makespan gaps w.r.t. training epochs for different models. \textbf{(b)} Time complexity analysis: the total time cost w.r.t. testing job size. \textbf{(c)} Evaluation on job difference: the total makespan gaps w.r.t. the distribution deviation.}}
    \label{fig5} 
\end{figure*}

\noindent\textbf{4) Evaluations.} We evaluate these methods from three perspectives: the \textit{average} makespan, the \textit{average} makespan gap towards the expert method (the NEH algorithm), and the \textit{sum} of runtime. For all of the evaluation criteria, the lower the better. We run all the experiments for \textit{three} different seeds and record the \textit{average} results. We find that the results of our method are robust against different seeds.

\subsection{5.2. Results and analysis}
\label{ra}

\noindent\textbf{1) Comparative study.}  Table \ref{tab1} reveals the comparisons on the random datasets under the Gamma distribution. Table \ref{bench} shows the comparisons of the two benchmarks. 
According to Table \ref{tab1}, the NEH algorithm generally achieves the lowest makespan performance among the four heuristics. However, as job size increases, the time of NEH is rapidly increasing. Compared with the heuristics, learning methods are fast in computations when testing. The testing time of the RL model is a bit faster than ours. The reason is that the RL model merely uses a single network when testing, though it requires bi-network during training, and our IL model has the complex graph structure GGCN. However, our proposed IL model outperforms the RL model in the makespan accuracy, where the makespan gap towards the NEH algorithms decreases by 5.5\% on average from 6.8\% (4.9\%$\sim$8.7\%) to 1.3\% (0.4\%$\sim$3.4\%). According to Table \ref{bench}, our IL model also shows excellent generalization capability to the unseen benchmark instances, outperforming the RL model in the makespan accuracy, though a little bit slower in time.



\begin{table}
    \centering
    \setlength{\abovecaptionskip}{0cm}
\caption{Ablation study on different training models.} 
\label{ablation}
\resizebox{8.5cm}{!}{
\begin{tabular}{lcccccc}
    \toprule[2pt]
    \qquad Model & \multicolumn{2}{c}{Training} & \multicolumn{1}{c}{Testing} \\
    Training Method (Encoder) & No. Para. & Time per epoch & Makespan \\
    \midrule
    RL(LSTM) & 709k+267k &31.0s &  31.6       \\
    RL(GGCN) & 365k+267k &18.6s &  30.6       \\
    IL (LSTM) & 709k & 13.6s & 31.0            \\
    \textbf{IL(GGCN) [Ours]} & \textbf{365k} & \textbf{12.4s} & \textbf{30.2}   \\
    \bottomrule[2pt]
\end{tabular}}
\end{table}


\noindent\textbf{2) Ablation study.} We conduct the ablation study to show the efficacy of the GGCN encoder and IL method, as shown in Table \ref{ablation} and Figure \ref{fig5}(a). Pan et al. \cite{pan2020solving} used RL for training and LSTM for the encoder. Our method uses IL and GGCN. Both methods use the attention mechanism for the decoder. For different methods, we use the pre-trained model with $\hat{n}{=}20$ to validate and test on $1000$ instances with ${n}{=}20$. The results show that our IL model is much lighter with fewer parameters and trains much faster than others. Notably, compared with \cite{pan2020solving}, our model's network parameters are only 37\% of theirs.
Figure \ref{fig5}(a) reveals the training processes, where our model leads to faster and more stable convergence. 

\noindent\textbf{3) Time complexity analysis.} We give detailed time complexity analysis among our IL model and the state-of-the-art NEH heuristic \cite{sharma2021improved} and the state-of-the-art RL method \cite{pan2020solving}. The total complexity of our IL method consists of three parts: getting expert solutions, network training, and testing. Their complexities are summarized in Table \ref{tab5} and Figure \ref{fig5}(b). We use $F_{RL}, B_{RL}$ to denote the complexity of one forward pass and one backward pass of the RL network, while $F_{IL}, B_{IL}$ is for our IL method. With respect to the testing job size $n$, NEH has \textit{polynomial} cost while our model has merely \textit{linear} cost. From the mathematical perspective, with the testing job size $n$ and the number of testing instances $N_3$ increasing, the efficiency of the learning method will be further highlighted.

\noindent\textbf{4) Time cost analysis.} We use the same pre-trained model on $12800$ training instances with $\hat{n}{=}20,m{=}5$, then choose 5 different testing sets and separately analyze their total time costs, as shown in Table \ref{total}. For the NEH algorithm, the total time is just the testing time. For the RL method, the total time contains the training time and the testing time. For our IL method, the total time contains all three parts. From Table \ref{total}, we can see that for the sets with small job sizes such as (10,60,5), (10,100,5), and (100,20,5), the NEH is efficient. For the three generated sets, with the increasing of job size $n$, the total time of the NEH increases rapidly while the total time of learning-based methods go increasing steadily. Generally, our IL method has lower total time costs than the RL method, which corresponds to Figure \ref{fig5}(b).

\noindent\textbf{5) Evaluations on job differences.} As mentioned in Section 4.2.(1), we believe that the difference between job features could impact the final results. Then we follow the Normal distribution, set the mean $\mu$=6, set the standard deviation $\sigma$ from 0 to 6 for the testing sets, and evaluate how the makespan gaps towards NEH present for non-graph structure (LSTM) and graph structure (GGCN), when both using the IL method. As shown in Figure \ref{fig5}(c), when $\sigma$=0, all jobs are the same and both models show the same performance; When $\sigma$ increases, the advances of GGCN over LSTM firstly increase and then decrease. The graph structure generally outperforms the non-graph structure.

\noindent\textbf{6) More experiments are in the Appendix.} Due to the space limits, more experiments are put in the Appendix, including the evaluations on the Normal distribution (\textit{Appendix D}), the analysis of different machine numbers (\textit{Appendix E}), the composition analysis of GGCN (\textit{Appendix F}), and the evaluations on TSP (\textit{Appendix G}).

\section{Conclusion}
The PFSS is a significant CO problem with wide applications. Learning-based methods have gained increasing attention in recent years, however, the literature heavily relies on the reinforcement learning method. In this paper, we introduce why and how to use the graph-based imitation learning method in solving PFSS, which leads to a lighter network, faster and more stable convergence, and lower makespan gaps, compared to the state-of-the-art RL method. The extensive experiments on generated and benchmark datasets clearly exhibit the competitiveness of our proposed method. 






%% file: table.tex
\begin{table*}[t!]
    \setlength{\abovecaptionskip}{0cm}
    \centering 
	\caption{{Comparative study on the generated datasets. We compare our IL method with four mostly-used heuristics and the state-of-the-art RL method. $m$ is set to 5 for all phases. We use the pre-trained model with $\hat{n}{=}20$, to test $1000$ instances with $n{=}20,50,100$ and test $100$ instances with $n{=}200,500,1000$, respectively. We report the \textit{average} makespan, the \textit{average} makespan gap, and the \textit{sum} of runtime. All results are the average of \textit{three} trials. The best performances are in \textbf{BOLD} among the heuristics and the learning methods, respectively. '$\bullet$' indicates that the makespan decrease of our method over the baseline method is statistically significant (via Wilcoxon signed-rank test at 5\% significance level.)}}
	\label{tab1}
	{\small
		\resizebox{17.5cm}{!}{
	\begin{tabular}{ccccccccccc}
	\toprule[1.5pt]
    \multicolumn{1}{c}{\quad} & \multicolumn{1}{c}{Job Size for Testing} & \multicolumn{3}{c}{PFSS-20} &  \multicolumn{3}{c}{PFSS-50} & \multicolumn{3}{c}{PFSS-100}  \\
    \midrule 
    Type & \multicolumn{1}{c}{Method} & Makespan$\downarrow$ & Gap$\downarrow$  & Time$\downarrow$ & Makespan$\downarrow$ & Gap$\downarrow$  & Time$\downarrow$ & Makespan$\downarrow$ & Gap$\downarrow$ & Time$\downarrow$ \\
    \cmidrule(lr){1-2}\cmidrule(lr){3-5}\cmidrule(lr){6-8}\cmidrule(lr){9-11}
    \multirow{4}{*}{Heuristics} & {Random search \cite{zabinsky2009random}}  & 34.1 $\bullet$ & 16.8\% & \textbf{1.2s} & 72.5 $\bullet$ & 16.0\% & \textbf{2.8s} & 132.4 $\bullet$ & 12.8\% &\textbf{5.4s}  \\
     &{Iterated local search \cite{lourencco2019iterated}}  & 30.0 & 2.7\% & 31.0s & 65.2 $\bullet$ & 4.3\% & 69.7s & 121.7 $\bullet$ & 3.7\% & 131.2s   \\
     &{Iterated greedy \cite{ruiz2019iterated}} & {29.3} &  0.3\% & 215.9s & 63.5 & 1.6\% & 1303.9s & 119.2 & 1.5\% & 5207.9s  \\
    &{NEH \cite{sharma2021improved} [Expert]}  &  \textbf{29.2} & \textbf{0.0\%} & {12.9s} & \textbf{62.5} & \textbf{0.0\%} & {200.1s} & \textbf{117.4} & \textbf{0.0\%} & {1544.6s}  \\

    \cmidrule(lr){1-2}\cmidrule(lr){3-5}\cmidrule(lr){6-8}\cmidrule(lr){9-11}
    RL & Actor-critic \cite{pan2020solving}  & 31.6 $\bullet$ & 8.2\% & \textbf{3.8s} & 66.8 $\bullet$ & 6.9\% & \textbf{4.5s} & 123.8 $\bullet$ & 5.4\% & \textbf{6.3s} \\
    IL & {Behavioral cloning [Ours]} & \textbf{30.2} & \textbf{3.4\%} &{4.0s} & \textbf{63.7} & \textbf{1.9\%} & {5.1s} & \textbf{118.3} & \textbf{0.8\%} & {7.2s} \\


    \midrule[1.5pt]
    \multicolumn{1}{c}{\quad} & \multicolumn{1}{c}{Job Size for Testing} & \multicolumn{3}{c}{PFSS-200} &  \multicolumn{3}{c}{PFSS-500} & \multicolumn{3}{c}{PFSS-1000} \\
    \midrule
    Type & \multicolumn{1}{c}{Method} & Makespan$\downarrow$ & Gap$\downarrow$  & Time$\downarrow$ & Makespan$\downarrow$ & Gap$\downarrow$  & Time$\downarrow$ & Makespan$\downarrow$ & Gap$\downarrow$ & Time$\downarrow$ \\
    \cmidrule(lr){1-2}\cmidrule(lr){3-5}\cmidrule(lr){6-8}\cmidrule(lr){9-11}
    \multirow{4}{*}{Heuristics} & {Random search \cite{zabinsky2009random}}  & 243.8 $\bullet$ & 9.0\% & \textbf{1.1s} & 570.5 $\bullet$ & 6.2\% & \textbf{2.7s} & 1103.4 $\bullet$ & 5.3\% & \textbf{54.2s}  \\
     &{Iterated local search \cite{lourencco2019iterated}}  &  230.2 $\bullet$ & 3.0\% & 25.1s & 549.4 $\bullet$ & 2.5\% &  65.4s & 1073.1 $\bullet$ & 2.4\% & 140.0s    \\

     &{Iterated greedy \cite{ruiz2019iterated}} & 229.3 $\bullet$  & 2.5\% & 0.6h & 542.1 & 0.9\% & 3.7h & 1060.8 & 1.2\% & 14.6h  \\
    &{NEH \cite{sharma2021improved} [Expert]} & \textbf{223.6} & \textbf{0.0\%} & {0.3h} & \textbf{537.2} & \textbf{0.0\%} & {5.5h} & \textbf{1048.2} & \textbf{0.0\%} & {43.1h}  \\ 
    \cmidrule(lr){1-2}\cmidrule(lr){3-5}\cmidrule(lr){6-8}\cmidrule(lr){9-11}
    RL & Actor-critic \cite{pan2020solving}  & 234.6 $\bullet$ & 4.9\%  & \textbf{17.2s} & 574.2 $\bullet$ & 6.9\% & \textbf{41.4s} & 1138.9 $\bullet$ & 8.7\% & \textbf{81.8s} \\
    IL & {Behavioral cloning [Ours]} & \textbf{224.6} & \textbf{0.4\%} &{20.0s} & \textbf{540.7} & \textbf{0.6\%} & {45.3s} & \textbf{1055.5} & \textbf{0.7\%} & {89.9s} \\
    \bottomrule[1.5pt]
	\end{tabular}}}
\end{table*}

%% file: benchmark2.tex
\begin{table*}[t!]
\setlength{\abovecaptionskip}{0cm}
\centering 
\caption{Comparative study on the benchmarks Taillard and VRF. We compare our IL method with NEH heuristics and the state-of-the-art RL method. The better performances between ours and the RL method are in \textbf{BOLD}. Refer to Subsection "Implementation details" below for more details of the setup.}  \label{bench}
\resizebox{17.5cm}{!}{
\begin{tabular}{cccccccccccc} 
	\toprule[1.5pt]
    & Instance size for testing: ($n$,$m$) & \multicolumn{2}{c}{(50,5)} &  \multicolumn{2}{c}{(100,5)} & \multicolumn{2}{c}{(100,20)} & \multicolumn{2}{c}{(200,20)} & \multicolumn{2}{c}{(500,20)}  \\
    \midrule 
    \multirow{4}{*}{Taillard \cite{taillard1993benchmarks}} & Method & Gap $\downarrow$  & Time$\downarrow$ & Gap$\downarrow$  & Time$\downarrow$ & Gap$\downarrow$  & Time$\downarrow$ & Gap$\downarrow$  & Time$\downarrow$ & Gap$\downarrow$ & Time$\downarrow$\\
    \cmidrule(lr){2-4}\cmidrule(lr){5-6}\cmidrule(lr){7-8}\cmidrule(lr){9-10}\cmidrule(lr){11-12} 
    & NEH \cite{sharma2021improved} & {0\%} & {2.0s} & {0\%} & 15.9s & {0\%} & 58.8s & {0\%} & 468.2s & {0\%} & 2.0h  \\
    & RL \cite{pan2020solving} & 14.6\% & \textbf{3.3s} & 13.2\% & \textbf{3.5s} &   \textbf{12.5\%} & \textbf{3.6s} & 12.6\% & \textbf{3.8s} & 9.9\% & \textbf{5.1s} \\
    & {IL [Ours]} & \textbf{12.8\%} & {4.5s} & \textbf{12.3\%} & {3.6s} & \textbf{12.5\%} & {4.8s} & \textbf{10.7\%} &4.6s & \textbf{9.4\%} & 5.2s \\
    \midrule[1.5pt]
    
    & Instance size for testing: ($n$,$m$) & \multicolumn{2}{c}{(40,5)} &  \multicolumn{2}{c}{(60,5)} & \multicolumn{2}{c}{(600,20)} & \multicolumn{2}{c}{(700,20)} & \multicolumn{2}{c}{(800,20)} \\ 
    \midrule 

    \multirow{4}{*}{VRF \cite{vallada2015new}} & Method & Gap $\downarrow$  & Time$\downarrow$ & Gap$\downarrow$  & Time$\downarrow$ & Gap$\downarrow$  & Time$\downarrow$ & Gap$\downarrow$  & Time$\downarrow$ & Gap$\downarrow$ & Time$\downarrow$\\
    \cmidrule(lr){2-4}\cmidrule(lr){5-6}\cmidrule(lr){7-8}\cmidrule(lr){9-10}\cmidrule(lr){11-12}
    & NEH \cite{sharma2021improved} & {0\%} & {0.9s} & {0\%} & 3.4s & {0\%} & 3.4h & {0\%} & 5.8h & {0\%} & 8.3h \\  
    & RL \cite{pan2020solving} & 17.4\%  & \textbf{3.5s} & 14.5\% & \textbf{3.3s} & 9.7\% & \textbf{4.2s} & 10.0\% & \textbf{5.3s} & 9.1\% & \textbf{5.1s}  \\
    & {IL [Ours]} & \textbf{15.9\%}  &{4.1s} & \textbf{14.1\%} & {4.8s} & \textbf{9.1\%} & 5.6s & \textbf{9.5\%} & 5.5s & \textbf{8.2\%} & 6.4s \\
    \bottomrule[1.5pt]
    
\end{tabular}}
\end{table*}

%% file: appendix.tex
\appendix 

\section*{Appendix}

\section{More details about related works}

In the Section "Related work" of the main paper, we have discussed the two categories of permutation flow shop scheduling (PFSS) methods, including the traditional-optimization-based methods and the learning-based methods, here we would like to discuss more regarding the learning-based methods for combinatorial optimization, and imitation learning.




\noindent\textbf{Machine learning for combinatorial optimization.} \ The application of machine learning (ML) to combinatorial optimization has been a popular topic with various approaches in the literature \cite{bengio2020machine}. Andrychowicz et al. \cite{andrychowicz2016learning} and Li et al. \cite{li2016learning} propose the idea of learning to optimize, where neural networks are used as the optimizer to minimize some loss functions. Pointer network (PN) \cite{vinyals2015pointer} is introduced as a model that uses attention to output a permutation of the input. Nazari et al. \cite{nazari2018reinforcement} replace the LSTM encoder of the PN by element-wise projections. Dai et al. \cite{dai2017learning} do not use a separate encoder and decoder, but a single model based on graph embeddings.  Nowak et al. \cite{nowak2017note} train a Graph Neural Network (GNN) in a supervised manner to directly output a tour as an adjacency matrix. Kool et al. \cite{kool2018attention} propose a model based on attention layers \cite{vaswani2017attention} to solve the problems. Gasse el al. propose a GNN model for learning branch-and-bound policies \cite{gasse2019exact}.

    
\noindent\textbf{Imitation learning.} \ Imitation learning (IL) techniques aim to mimic the behavior of an expert or teacher in a given task \cite{hussein2017imitation}. IL and RL both work for the Markov decision processes (MDP). RL tends to have the agent learn from scratch through its exploration with a specified reward function, however, the agent of IL does not receive task reward but learns by observing and mimicking \cite{torabi2019recent}. Similar to traditional supervised learning where the samples represent pairs of features and ground-truth labels, IL has the samples demonstrating pairs of states and actions. Broadly speaking, research in the IL can be split into two main categories: behavioral cloning (BC) \cite{torabi2018behavioral}, and inverse reinforcement learning (IRL) \cite{abbeel2004apprenticeship}. In this paper, we choose BC for the study.

\section{Mathematical formulations}
\label{problem}
In the PFSS problem, we denote the number of jobs as $n$ and machines as $m$. Each job is a compromise of $m$ operations, and one operation corresponds to only one machine. All jobs share the same processing order when passing through the machines. Operations can not be interrupted, and one machine can only process one job at a time. The goal of the PFSS problem is to find the best job permutation which minimizes the makespan. We give the following notations for the mathematical formulations of the PFSS problem,  

\begin{itemize}
    \item Indices and sets: 
    \begin{itemize}
        \item $i$: machine index; $i \in I$=$\{1,2,...,m\}$; 
        \item $j,k,l$: job index; $j,k,l \in J$=$\{1,2,...,n\}$;  
        \item $J'$=$J{\cup}\{0\}$: extended set of jobs with dummy job 0;
    \end{itemize}

    \item Inputs:  
    \begin{itemize}
        \item $x_{ij}$: the processing times of job $j$ on machine $i$; $x_{ij} {\geq} 0, \  x_{ij} {\in} X_{m\times n}$;
    \end{itemize}

    \item Common variables:
    \begin{itemize}
        \item $A_{i}$: the upper bound on the time at which machine $i$ finishes processing its last job; $A_0$=0, $A_i$=$A_{i-1}{+}\sum_{j\in J}{x_{ij}}$
        \item $y_{ij}$: the starting time of job $j$ on machine $i$, a non-negative real variable;
        \item $C_{max}$: the makespan (the total processing time of all jobs, also known as the finishing time of the last job on the last machine), a non-negative real variable;
        \item $z_{jk}$: the job precedence; $j{\neq}k$; $z_{jk}$=$1$ denotes that job $j$ is the immediate predecessor of job $k$ and otherwise for $z_{jk}$=$0$. Notice that $z_{0j}$=$1$ implies job $j$ is the first job of the permutation and $z_{j0}$=$1$ means job $j$ is the last job of the permutation. 
    \end{itemize}

    \item Outputs:
    \begin{itemize}
        \item $\tau^{*}$=$[\tau_0^{*},\tau_1^{*},...,\tau_{n-1}^{*}]$: the optimal permutation determined by the job precedence $z_{jk}$, where the makespan $C_{max}$ is minimized. 
    \end{itemize}
\end{itemize}

According to \cite{rios1998computational}, the PFSS problem is formulated as a mixed-integer programming model, 

\begin{align}
    \mathop{\min} \ C_{max} \qquad\qquad\qquad\quad \label{eq1_app}  \\
    s.t. \ \ \ \ \ \ \ 
        \sum_{j{\in}J_0}^n z_{jk} = 1, k{\in}J_0, \qquad\qquad\quad \label{eq2_app} \\
        \sum_{k{\in}J_0}^n z_{jk} = 1, j{\in}J_0, \qquad\qquad\quad \label{eq3_app} \\
        y_{ij} + x_{ij} \leq y_{ik} + A_i(1-z_{jk}), i \in I, j,k\in J, \label{eq4_app} \\
        y_{mj} + x_{mj} \leq C_{max}, j \in J, \qquad\qquad \label{eq5_app} \\
        y_{ij} + x_{ij} \leq y_{i{+}1,j}, i \in I\backslash\{m\}, j\in J, \qquad \label{eq6_app} \\
        z_{jk} \in \{0,1\}, j,k\in J_0, j \neq k, \qquad\quad \label{eq7_app} \\
        y_{ij} \geq 0, i \in I, j \in J. \qquad\qquad\quad \label{eq8_app}
\end{align}

Equations \eqref{eq2_app} and \eqref{eq3_app} imply that every job must have one predecessor and successor, respectively, and one machine can only process one job at one time. Condition \eqref{eq4_app} states that if job $j$ precedes job $k$, then the starting time of job $k$ on machine $i$ must not exceed the completion time of job $j$ on machine $i$, which is a set of sub-tour elimination constraints. Condition \eqref{eq5_app} assures that the makespan is greater than or equal to the completion time of all jobs on the last machine. In contrast, condition \eqref{eq6_app} states that a job cannot start processing on one machine if it has not finished processing on the previous one. Conditions \eqref{eq7_app} and \eqref{eq8_app} define the domain of the variables. Importantly, we allow the possibility of processing times $x_{ij}=0$, meaning the PFSS problem with jobs missing in certain machines, which extends our PFSS models to wider applicable scenarios.

\begin{table}
    \centering
    \caption{Performance evaluations between the mathematical models and the NEH algorithm. We report the \textit{average} makespan and the \textit{sum} of runtime.}
    \label{tab1_appendix}
    \resizebox{8cm}{!}{
    \begin{tabular}{cccccc}
    \toprule[2pt]
    \multicolumn{2}{c}{Size} & \multicolumn{2}{c}{Math. Model} & \multicolumn{2}{c}{NEH Algorithm}\\ 
    No. of Job & No. of Machine & Makespan & Time & Makespan & Time \\
    \cmidrule(lr){1-2}\cmidrule(lr){3-4}\cmidrule(lr){5-6}
    8 & 5 & \textbf{16.8} & 309.72s & 17.7 & \textbf{0.09s}      \\
    9 & 5 & \textbf{17.2} & 1867.17s & 18.0 & \textbf{0.12s}      \\
    10 & 5 & \textbf{17.9} & 11394.49s & 18.2 & \textbf{0.16s}     \\  
    \bottomrule[2pt]
    \end{tabular}}
\end{table}

\section{Expert for imitation learning: NEH}
Mathematical models are exact methods which is guaranteed to obtain the optimal solutions, and it is naturally ideal to use the mathematical models as the expert method. We thus construct the mixed-integer programming model as formulated in Equations (1)-(8) according to \cite{rios1998computational} through the commercial software Gurobi \cite{bixby2007gurobi}.  

We show the performance comparisons between the mathematical model \cite{rios1998computational} and the NEH algorithm \cite{sharma2021improved} as one of the state-of-the-art heuristics. For each method, we randomly generate $100$ instances following Gamma distribution for testing. For each task, we evaluate the makespan and runtime of different methods by recording the average makespan and the sum of runtime. As shown in Table \ref{tab1_appendix}, the computational time of the mathematical model is much slower than the NEH algorithm, while their makespan gaps are not much too wide. 
Moreover, some of the current popular exact methods for the PFSS problem, such as \cite{gmys2020computationally}, require that the processing times of inputs only to be integers, while our randomly generated instances are all real numbers. 

{Considering the great amount of training and testing instances, even some of them are large-scale, we decide to choose the state-of-the-art heuristics NEH algorithm \cite{sharma2021improved} instead of the exact mathematical models as the expert method.}

\section{Evaluations on normal distribution}

\input{appendix/table}

\noindent\textbf{Setup.} We also show the experimental results regarding the normal distribution with $\mu{=}6, \sigma{=}6$ in this section. We pre-process the datasets by setting all the negative elements of datasets to zeros. We also find that the training of our method is stable and the results are robust against different seeds in this normal distribution. We follow the same setting as the experiments on Gamma distribution. Let $N_1, N_2, N_3$ denote the number of training, validation, and testing instances, and $\hat{n}, n, m$ denote the training job size, the testing job size, and the machine size, respectively. $m$ is set to 5 for all phases. We train the model on $N_1{=}12800$ instances with $\hat{n}{=}20$, validate on $N_2{=}1000$ instances with the same job size. We then generalize the small pre-trained model to any larger job sizes. We test on $N_3{=}1000$ instances with $n{=}20,50,100$, and test on $N_3{=}100$ instances with $n{=}200,500,1000$, respectively.         

\begin{figure}[t] 
    \centering 
    \includegraphics[width=0.33\textwidth]{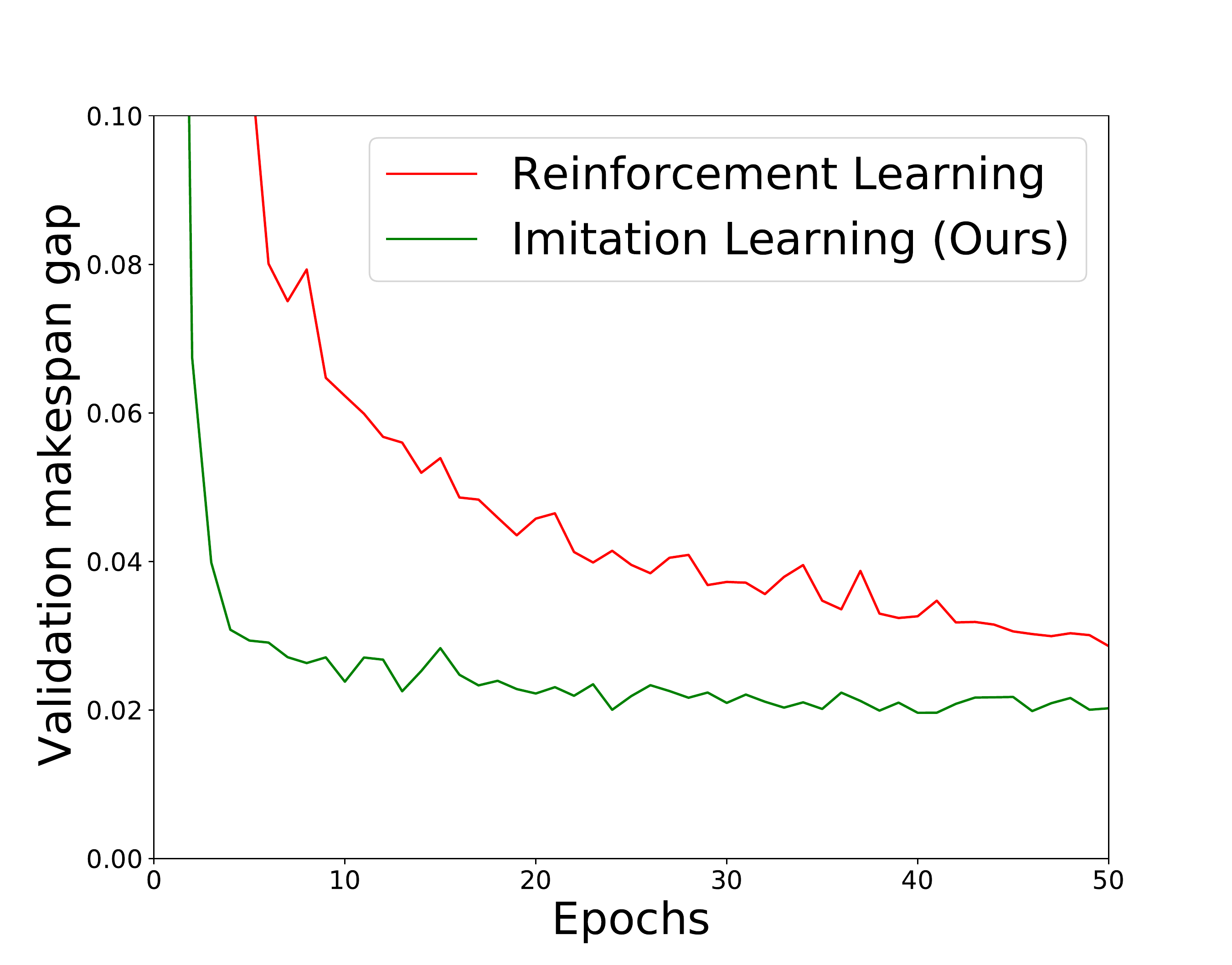}
    \caption{Comparison study on training convergence: we compare our imitation learning model with the state-of-the-art reinforcement learning model during the training under the normal distribution. The vertical axis shows the validation makespan gap towards the expert method (NEH).}
    \label{fig1_appendix} 
\end{figure}

\noindent\textbf{Training analysis}. We compare the state-of-the-art reinforcement learning (RL) model with our proposed imitation learning (IL) model under the normal distribution dataset. The state-of-the-art RL model has a bi-network structure with an actor-network and a critic network, possessing more network parameters than the IL model and taking more time in training. The RL model utilizes the LSTM module as encoder whereas our IL model chooses GGCN. Figure \ref{fig1_appendix} reveals the training performances of the two different models. By applying the IL training paradigm, the validation makespan decreases compared to that using the RL paradigm. And it is evident that IL models converge faster and more stable. Unlike RL models, which explore their paths slowly during the training, IL models start to jump the makespan gap at early stage. The IL model almost converges around the 10-20 epochs, while the RL model explores until the convergence around 40-50 epochs. Consequently, our proposed IL model can efficiently map the represented features by GGCN directly to the actions guided by the expert method, leading to a more accurate makespan and a faster training process. With the combination of the GGCN and the IL paradigm, we can obtain a more accurate and even lighter model compared with the state-of-the-art RL model \cite{pan2020solving}.

\noindent\textbf{Comparison with the baselines}.  From Table \ref{tab2}, we can see that among the four heuristics, random search has shown the worst performance, though it is exceedingly fast in runtime. Iterated local search improves the makespan accuracy compared to the random search, while it takes up a lot more runtime in computation. Iterated greedy costs tremendously more time than iterated local search due to the repeated construction and destruction processes, and it is indeed working better than the iterated local search. NEH algorithm generally achieves the best performance among the four heuristics, obtaining the lowest makespan. Compared to iterated greedy, the NEH algorithm achieves a more accurate makespan and computes more efficiently when the job number is no more than 200. When the job number exceeds 200, the NEH algorithm computes for the longest time, even though it achieves the best makespan performance. Compared to the RL model, our IL model overall obtains a better performance.

\input{appendix/machinenum}

\begin{table}[!t]
    \centering
    \caption{GGCN composition analysis on aggregation functions.}
    \label{tab4_appendix}
    \resizebox{8cm}{!}{
    \begin{tabular}{ccccccccc}
    \toprule[2pt]
    \multicolumn{1}{c}{\quad} & \multicolumn{2}{c}{PFSS-20} & \multicolumn{2}{c}{PFSS-50} & \multicolumn{2}{c}{PFSS-100} \\
    Method & Makespan & Time & Makespan & Time & Makespan & Time \\
    \midrule
    MAX & 36.4 & 4.0s & 76.6 & 5.9s & 139.9 & 7.3s    \\
    SUM & 34.3 & 4.2s & 72.1 & 5.3s & 133.7 & 7.7s      \\
    \textbf{MEAN} & \textbf{30.2} & \textbf{4.0s} & \textbf{63.7} & \textbf{5.1s} & \textbf{118.3} & \textbf{7.2s}  \\
    \bottomrule[2pt]
    \end{tabular}}
\end{table}

\begin{table}[!t]
    \centering
    \caption{GGCN composition analysis on normalization styles.}
    \label{tab5_appendix}
\resizebox{8cm}{!}{
\begin{tabular}{ccccccccc}
    \toprule[2pt]
    \multicolumn{1}{c}{\quad} & \multicolumn{2}{c}{PFSS-20} & \multicolumn{2}{c}{PFSS-50} & \multicolumn{2}{c}{PFSS-100} \\
    Method & Makespan & Time & Makespan & Time & Makespan & Time \\
    \midrule
    None & 30.4 & 4.6s & 64.1 & 5.6s & 119.1 & 7.4s \\ 
    Layer   & 30.3 & 4.7s & 63.9 & 6.0s & 118.7 & 8.1s      \\
    \textbf{Batch} & \textbf{30.2} & \textbf{4.0s} & \textbf{63.7} & \textbf{5.1s} & \textbf{118.3} &  \textbf{7.2s}       \\
    \bottomrule[2pt]
\end{tabular}}
\end{table}

\section{Evaluations on machine number}

We evaluate how different machine numbers $m$ will impact the performance. We extend the machine number $m$ to 10, keep all other parameters and settings the same as the experiments in the paper, and record the experimental results. When $m$ is set to 10, our proposed IL model outperforms the RL model in the makespan accuracy, where the makespan gap towards the NEH algorithms decreases by 6.5\% on average from 11.1\% (3.7\%$\sim$17.4\%) to 4.6\% (0.8\%$\sim$8.5\%). 

From the results, as shown in Table \ref{tab3}, we can see that a larger machine number $m$ may probably achieve larger average makespan gaps. Compared to average gap decrease 5.5\% for $m{=}5$, a larger $m$ ($m{=}10$) has larger average gap decrease, which is 6.5\%.

\section{Composition analysis of GGCN}

The component selection of GGCN will affect the feature extraction ability of the model. We evaluate the effectiveness of different aggregation functions and different normalization styles. We use the pre-trained model with $\hat{n}{=}20$ to test on $1000$ instances with ${n}{=}20,50,100$, for both experiments. 

We mainly consider three aggregation functions for graph aggregation styles, namely MAX, SUM, and MEAN. Table \ref{tab4_appendix} shows that MEAN aggregation generally outperforms the others. For normalization styles, we consider no normalization, layer normalization \cite{ba2016layer}, and batch normalization \cite{ioffe2015batch}. 

Table \ref{tab5_appendix} reveals that the batch normalization performs better than the other two. According to Table \ref{tab4_appendix} and Table \ref{tab5_appendix}, it can be found that the composition of different GGCN has a great influence on the makespan, which also proves that a suitable GGCN can provide a better context embedding for the decoding process.

\section{Evaluations on other permutation problems} 

 We evaluate how our models are applied to other permutation problems. We show the results using our model for the traveling salesman problem (TSP) \cite{reinelt2003traveling}. TSP can be seen as a special case of permutation problem. 
 
 Detailed results are given in Table \ref{tsp}. We compare our IL method with the exact solver Gurobi \cite{bixby2007gurobi} and the RL method \cite{pan2020solving}. We report the average tour length gap towards Gurobi and the sum of runtime. For our IL method, we simply regard the Gurobi result as the expert. From the results, IL method presents smaller gaps than RL, though IL costs a little bit more runtime.

\begin{table}[t]
 \centering 
 	\caption{Evaluations on TSP problem using different methods: we compare our IL method with the RL method and exact solver. The best performance between our IL method and RL method is in \textbf{BOLD}.}
 	\label{tsp}
	{\small
		\resizebox{8.5cm}{!}{
	\begin{tabular}{ccccccc}
	\toprule[1.5pt]
   Size: ($N,n,m$) & \multicolumn{2}{c}{(1000,20,2)} &  \multicolumn{2}{c}{(1000,50,2)} & \multicolumn{2}{c}{(1000,100,2)}  \\
    \midrule
    Method & Gap$\downarrow$ & Time$\downarrow$  & Gap$\downarrow$   & Time$\downarrow$  & Gap$\downarrow$   & Time$\downarrow$  \\
    \cmidrule(lr){1-3}\cmidrule(lr){4-5}\cmidrule(lr){6-7}
    Gurobi  & {0\%} & 13.9s & {0\%} & 155.9s & {0\%} & 1355.0s \\
    RL   & 1.4\% & \textbf{3.9s} & 5.2\% & \textbf{5.8s} &  13.1\% & \textbf{6.1s} \\
    {IL} & \textbf{0.7\%} & {4.2s} & \textbf{3.6\%} & {6.6s} & \textbf{8.3\%} & {7.9s}\\
	\bottomrule[1.5pt]
	\end{tabular}}}
\end{table} 

%% file: appendix/table.tex
\begin{table*}[!t]
	\centering 
	\caption{Comparative study on Normal distribution when machine number $m$ is set to 5: we compare our IL model (GGCN as encoder) with four mostly-used heuristics and the state-of-the-art RL method (LSTM as encoder). We use the pre-trained model with $\hat{n}{=}20$, and test on $1000$ instances with $n{=}20,50,100$ and test on $100$ instances with $n{=}200,500,1000$ via generalization. We report the \textit{average} makespan, the \textit{average} makespan gap, and the \textit{sum} of runtime. For all evaluation criteria, the \textit{lower} the better. All results are the average of \textit{three} trials. The best performances are in \textbf{BOLD} among the heuristics and the learning-based methods, respectively. '$\bullet$' indicates that the makespan decrease of our method over the baseline method is statistically significant (via Wilcoxon signed-rank test at 5\% significance level.)}
	\label{tab2}
	{\small
		\resizebox{18cm}{!}{
	\begin{tabular}{ccccccccccc}
	\toprule[1.5pt]
    \multicolumn{1}{c}{\quad} & \multicolumn{1}{c}{Job Size for Testing} & \multicolumn{3}{c}{PFSS-20} &  \multicolumn{3}{c}{PFSS-50} & \multicolumn{3}{c}{PFSS-100}  \\
    \midrule 
    Type & \multicolumn{1}{c}{Method} & Makespan$\downarrow$ & Gap$\downarrow$  & Time$\downarrow$ & Makespan$\downarrow$ & Gap$\downarrow$  & Time$\downarrow$ & Makespan$\downarrow$ & Gap$\downarrow$ & Time$\downarrow$ \\
    \cmidrule(lr){1-2}\cmidrule(lr){3-5}\cmidrule(lr){6-8}\cmidrule(lr){9-11}
    \multirow{4}{*}{Heurisctics} & {Random search \cite{zabinsky2009random}}  & 197.4 $\bullet$ & 16.0\% & \textbf{1.1s} & 424.2 $\bullet$ & 12.5\% & \textbf{3.7s} & 783.9 $\bullet$ & 9.4\% &\textbf{5.6s}  \\
     &{Iterated local search \cite{lourencco2019iterated}}  & 175.4 $\bullet$ & 3.1\% & 32.5s & 393.0 $\bullet$ & 4.2\% & 68.1s & 740.3 $\bullet$ & 3.4\% & 126.6s   \\
     &{Iterated greedy \cite{ruiz2019iterated}} & {170.5} &  0.2\% & 202.5s & 381.4 & 1.1\% & 1432.8s & 722.9 $\bullet$ & 1.0\% & 5267.8s  \\
    &{NEH \cite{sharma2021improved} [Expert]}  &  \textbf{170.1} & \textbf{0.0\%} & {12.9s} & \textbf{377.2} & \textbf{0.0\%} & {200.1s} & \textbf{716.0} & \textbf{0.0\%} & {1473.2s}  \\

    \cmidrule(lr){1-2}\cmidrule(lr){3-5}\cmidrule(lr){6-8}\cmidrule(lr){9-11}
    RL & Actor-critic \cite{pan2020solving}  & 175.3 $\bullet$ & 3.3\% & \textbf{3.5s} & 381.7 $\bullet$ & 1.3\% & \textbf{5.5s} & 722.8 $\bullet$ & 0.8\% & \textbf{6.9s} \\
    IL & {Behavioral cloning (Ours)} & \textbf{172.9} & \textbf{2.1\%} &{4.1s} & \textbf{379.7} & \textbf{0.5\%} & {5.8s} & \textbf{718.8} & \textbf{0.2\%} & {7.6s} \\

    \midrule[1.5pt]
    \multicolumn{1}{c}{\quad} & \multicolumn{1}{c}{Job Size for Testing} & \multicolumn{3}{c}{PFSS-200} &  \multicolumn{3}{c}{PFSS-500} & \multicolumn{3}{c}{PFSS-1000} \\
    \midrule
    Type & \multicolumn{1}{c}{Method} & Makespan$\downarrow$ & Gap$\downarrow$  & Time$\downarrow$ & Makespan$\downarrow$ & Gap$\downarrow$  & Time$\downarrow$ & Makespan$\downarrow$ & Gap$\downarrow$ & Time$\downarrow$ \\
    \cmidrule(lr){1-2}\cmidrule(lr){3-5}\cmidrule(lr){6-8}\cmidrule(lr){9-11}
    \multirow{4}{*}{Heurisctics} & {Random search \cite{zabinsky2009random}}  & 1486.9 $\bullet$ & 9.0\% & \textbf{1.0s} & 3586.4 $\bullet$ & 6.2\% & \textbf{2.1s} & 7024.5 $\bullet$ & 5.3\% & \textbf{53.3s}  \\
     &{Iterated local search \cite{lourencco2019iterated}}  &  1456.8 $\bullet$ & 3.0\% & 20.4s & 3485.3 $\bullet$ & 2.5\% &  65.4s & 6966.8 $\bullet$ & 2.4\% & 140.0s    \\

     &{Iterated greedy \cite{ruiz2019iterated}} & 1423.4 $\bullet$ & 2.5\% & 0.8h & 3450.9 & 0.9\% & 3.9h & 6927.6 $\bullet$ & 1.2\% & 13.9h  \\
    &{NEH \cite{sharma2021improved} [Expert]} & \textbf{1415.9} & \textbf{0.0\%} & {0.4h} & \textbf{3448.6} & \textbf{0.0\%} & {5.7h} & \textbf{6917.8} & \textbf{0.0\%} & {40.8h}  \\ 
    \cmidrule(lr){1-2}\cmidrule(lr){3-5}\cmidrule(lr){6-8}\cmidrule(lr){9-11}
    RL & Actor-critic \cite{pan2020solving}  & 1424.8 $\bullet$ & 0.6\%  & \textbf{19.4s} & 3479.8 $\bullet$ & 0.9\% & \textbf{43.5s} & 6955.2 $\bullet$ & 0.5\% & \textbf{79.5s} \\
    IL & {Behavioral cloning (Ours)} & \textbf{1418.7} & \textbf{0.2\%} &{21.7s} & \textbf{3457.2} & \textbf{0.3\%} & {50.4s} & \textbf{6920.7} & \textbf{0.1\%} & {88.4s} \\
    \bottomrule[1.5pt]
	\end{tabular}}}
\end{table*}

%% file: appendix/machinenum.tex
\begin{table*}[t]
	\caption{{Comparative study on the Gamma distribution when machine number $m$ is set to 10: we compare our IL method with four mostly-used heuristics and the state-of-the-art RL method. We use the pre-trained model with $\hat{n}{=}20$, test on $1000$ instances with $n{=}20,50,100$ and test on $100$ instances with $n{=}200,500,1000$ via generalization. We report the \textit{average} makespan, the \textit{average} makespan gap, and the \textit{sum} of runtime. For all evaluation criteria, the \textit{lower} the better. All results are the average of \textit{three} trials.}}
	\label{tab3}
	{\small
		\resizebox{17.5cm}{!}{
	\begin{tabular}{ccccccccccc}
	\toprule[1.5pt]
    \multicolumn{1}{c}{\quad} & \multicolumn{1}{c}{Job Size for Testing} & \multicolumn{3}{c}{PFSS-20} &  \multicolumn{3}{c}{PFSS-50} & \multicolumn{3}{c}{PFSS-100}  \\
    \midrule 
    Type & \multicolumn{1}{c}{Method} & Makespan & Gap  & Time & Makespan & Gap  & Time & Makespan & Gap  & Time \\
    \cmidrule(lr){1-2}\cmidrule(lr){3-5}\cmidrule(lr){6-8}\cmidrule(lr){9-11}
    \multirow{4}{*}{Heurisctics} & {Random search \cite{zabinsky2009random}}  & 48.2 & 25.2\% & \textbf{2.2s} & 94.0 & 33.1\% & \textbf{5.4s} & 161.5 & 28.8\% &\textbf{11.1s}  \\
     &{Iterated local search \cite{lourencco2019iterated}}  & 40.6 & 5.7\% & 59.7s & 79.9 & 13.2\% & 149.4s & 140.3 & 11.9\% & 325.1s   \\
     &{Iterated greedy \cite{ruiz2019iterated}} & {38.6} &  0.5\% & 382.3s & 71.9 & 1.8\% & 2430.0s & 132.0 & 5.3\% & 2.9h  \\
     
    &\textbf{NEH \cite{sharma2021improved} (Expert)}  &  \textbf{38.4} & \textbf{0.0\%} & {22.7s} & \textbf{70.6} & \textbf{0.0\%} & {334.3s} & \textbf{125.4} & \textbf{0.0\%} & {2623.1s}  \\

    \cmidrule(lr){1-2}\cmidrule(lr){3-5}\cmidrule(lr){6-8}\cmidrule(lr){9-11}
    RL & Actor-critic \cite{pan2020solving}  & 43.8 & 14.1\% & \textbf{3.9s} & 82.9 & 17.4\% & \textbf{5.1s} & 142.9 & 13.9\% & \textbf{6.4s} \\
    IL & \textbf{Behavioral cloning [Ours]} & \textbf{41.0} & \textbf{6.8\%} &{4.1s} & \textbf{76.6} & \textbf{8.5\%} & {5.9s} & \textbf{133.7} & \textbf{6.6\%} & {8.7s} \\

    \cmidrule(lr){1-2}\cmidrule(lr){3-5}\cmidrule(lr){6-8}\cmidrule(lr){9-11}
    \multicolumn{1}{c}{\quad} & \multicolumn{1}{c}{Job Size for Testing} & \multicolumn{3}{c}{PFSS-200} &  \multicolumn{3}{c}{PFSS-500} & \multicolumn{3}{c}{PFSS-1000} \\
    \midrule
    Type & \multicolumn{1}{c}{Method} & Makespan & Gap  & Time & Makespan & Gap  & Time & Makespan & Gap  & Time \\
    \cmidrule(lr){1-2}\cmidrule(lr){3-5}\cmidrule(lr){6-8}\cmidrule(lr){9-11}
    \multirow{4}{*}{Heurisctics} & {Random search \cite{zabinsky2009random}}  & 287.3 & 22.0\% & \textbf{2.1s} & 636.2 & 14.6\% & \textbf{5.2s} & 1184.4 & 6.8\% & \textbf{10.3s}  \\
     &{Iterated local search \cite{lourencco2019iterated}}  &  259.1 & 10.1\% & 64.5s & 587.3 & 5.8\% & 188.7s & 1149.2 & 3.7\% & 391.5s    \\

     &{Iterated greedy \cite{ruiz2019iterated}} & 240.8 & 2.3\% & 1.2h & 566.9 & 2.1\% & 7.3h & 1110.4 & 1.9\% & 29.2h  \\
    &\textbf{NEH \cite{sharma2021improved} (Expert)} & \textbf{235.4} & \textbf{0.0\%} & {0.6h} & \textbf{555.1} & \textbf{0.0\%} & {9.2h} & \textbf{1108.3} & \textbf{0.0\%} & {83.9h}  \\ 
    \cmidrule(lr){1-2}\cmidrule(lr){3-5}\cmidrule(lr){6-8}\cmidrule(lr){9-11}
    RL & Actor-critic \cite{pan2020solving}  & 260.1 & 10.5\%  & \textbf{26.6s} & 592.4 & 6.9\% & \textbf{59.0s} & 1148.9 & 3.7\% & \textbf{97.9.s} \\
    IL & \textbf{Behavioral cloning [Ours]} & \textbf{245.0} & \textbf{4.1\%} & {29.9s} & \textbf{561.5} & \textbf{1.1\%} & {61.1s} & \textbf{1118.9} & \textbf{0.8\%} & {108.6s} \\
    \bottomrule[1.5pt]
	\end{tabular}}}
\end{table*}